\documentclass[conference,a4paper]{ieeetran}

\usepackage[latin1]{inputenc}
\usepackage{subeqnarray, flushend}
\usepackage{amsfonts}
\usepackage{amsmath}
\usepackage{amssymb}
\usepackage{graphicx}
\usepackage{color}
\usepackage{cite}
\usepackage{tablefootnote}
\newcommand{\kimiapath}{\textit{Kimia Path24 }}

\usepackage{fancyheadings}
\title{Convolutional Neural Networks for Histopathology Image Classification: Training vs. Using Pre-Trained Networks}

\author{
	\authorblockN{
		Brady Kieffer\authorrefmark{1}, 
		Morteza Babaie\authorrefmark{2} 
		Shivam Kalra\authorrefmark{1}, 
		and H.R.Tizhoosh\authorrefmark{1} \vspace{0.1in}
	}
	\authorblockA{
		\authorrefmark{1} KIMIA Lab, University of Waterloo, ON, CANADA		
	}
	\authorblockA{
		\authorrefmark{2} Mathematics and Computer Science Department, Amirkabir University of Technology, Tehran, IRAN\\
		e-mails: bwkieffe@edu.uwaterloo.ca, M.Babaie@aut.ac.ir, shivam.kalra@uwaterloo.ca, tizhoosh@uwaterloo.ca
	}
}

\begin{document}

\maketitle

\begin{abstract}
We explore the problem of classification within a medical image data-set based on a feature vector extracted from the deepest layer of pre-trained Convolution Neural Networks. We have used feature vectors from several pre-trained structures, including networks with/without transfer learning to evaluate the performance of pre-trained deep features versus CNNs which have been trained by that specific dataset as well as the impact of transfer learning with a small number of samples. All experiments are done on \emph{Kimia Path24} dataset which consists of 27,055 histopathology training patches in 24 tissue texture classes along with 1,325 test patches for evaluation. The result shows that pre-trained networks are quite competitive against training from scratch. As well, fine-tuning does not seem to add any tangible improvement for VGG16 to justify additional training while we observed considerable improvement in retrieval and classification accuracy when we fine-tuned the Inception structure.
\end{abstract}

\begin{keywords}
Image retrieval, medical imaging, deep learning, CNNs,  digital pathology, image classification, deep features, VGG, Inception.
\end{keywords}

\section{Introduction}
We are amid a transition from traditional pathology to digital pathology where scanners are replacing microscopes rapidly. Capturing the tissue characteristics in digital formats opens new horizons for diagnosis in medicine. On on hand, we will need to store thousands and thousands of specimens in large physical archives of glass samples. This will be a relief for many hospitals with limited space. On the other hand, acquiring an image from the specimen enables more systematic analysis, collaborations possibilities and, last but not least, the computer-aided diagnosis for pathology, arguable the final frontier of vision-based disease diagnosis. However, like any other technology, digital pathology comes with its own challenges; whole-scan imaging generally generates gigapixel files that also require (digital) storage and are not easy to analyze via computer algorithms. Detection, segmentation, and identification of tissue types in huge digital images, e.g., 50,000$\times$70,000 pixels, appears to be a quite daunting task for computer vision algorithms. 

Looking at the computer vision community, the emergence of deep learning and its vast possibilities for recognition and classification seems to be a lucky coincidence when we intend to address the above-mentioned obstacles of digital pathology. Diverse deep architectures have been trained with large set of images, e.g., \emph{ImageNet} project or \emph{Faces in the Wild} database, to perform difficult tasks like object classification and face recognition. The results have been more than impressive; one may objectively speak of a computational revolution. Accuracy numbers in mid and high 90s have become quite common when deep networks, trained with millions of images, are tested to recognize unseen samples.              

In spite of all progress, one can observe that the applications of deep learning in digital pathology hast not fully started yet. The major obstacle appears to be the lack of large labelled datasets of histopathology scans to properly train some type of multi-layer neural networks, a requirement that may still be missing for some years to come. Hence, we have to start designing and training deep nets with the available datasets. Training from scratch when we artificially increase the number of images, i.e., data augmentation, is certainly the most obvious action. But we can also use nets that have been trained with millions of (non-medical) images to extract \emph{deep features}. As a last possibility, we could slightly train (fine-tune) the pre-trained nets to adjust them to the nature of or data before we use them as feature extractors or classifiers.   

In this paper, we investigate the usage of deep networks for \emph{Kimia Path24} via training from scratch, feature extraction, and fine-tuning. The results show that employing a pre-trained network (trained with non-medical images) may be the most viable option. 
 
\section{Background}
Over recent years researchers have shown interest in leveraging machine-learning techniques for digital pathology images. These images pose unique issues due to their high variation, rich structures, and large dimensionality. This has lead researchers to investigate various image analysis techniques and their application to digital pathology \cite{gurcan2009histopathological}. For dealing with the large rich structures within a scan, researchers have attempted segmentation on both local and global scales. For example, researchers have conducted works on the segmentation of various structures in breast histopathology images using methods such as thresholding, fuzzy c-means clustering, and adaptive thresholding with varying levels of success \cite{gurcan2009histopathological,naik2007gland,karvelis2006watershed,petushi2006large}. 

When applying these methods to histopathological images, it is often desired that a computer aided diagnosis (CAD) method be adopted for use in a content-based image retrieval (CBIR) system. Work has been done to propose various CBIR systems for CAD by multiple groups \cite{large-scale-retrieval}. Recently, hashing methods have been employed for large-scale image retrieval. Among the hashing methods, kernelized and supervised hashing are considered the most effective \cite{large-scale-retrieval,image-hashing}. More recently Radon barcodes have been investigated as a potential method for creating a CBIR \cite{tizhoosh2015barcode, tizhoosh2016minmax, morteza2017}.  

\textit{Yi et al.} utilized CNNs on a relatively small mammography dataset to achieve a classification accuracy of $ 85\% $ and an ROC AUC of $0.91$ whereas handcrafted features were only able to obtain an accuracy of $71\%$ \cite{breast-tumors}. 

Currently, there is interest in using pre-trained networks to accomplish a variety of tasks outside of the original domain \cite{survey-on-transfer-learning}. This is of great interest for medical tasks where there is often a lack of comprehensive labeled data to train a deep network \cite{cnn-transfer-learning}. Thus, other groups have leveraged networks trained on the ImageNet database which consists of more than 1.2 million categorized images of 1000+ classes \cite{deng2009imagenet,cnn-transfer-learning, cnn-transfer-learning-training}. These groups have reported a general success when attempting to utilize pre-trained networks for medical imaging tasks \cite{cnn-transfer-learning,cnn-transfer-learning-training,bar2015deep}. 

In this study we explore and evaluate the performance of a CNN when pre-trained on non-medical imaging data \cite{cnn-lecunn, cnn-transfer-learning}. Specifically, when used as feature extractors with and without fine tuning for a digital pathology task.

\section{Data Set}
The data used to train and test the CNNs was the \kimiapath  consisting of 24 whole scan images (WSIs), manually selected from more than 350 scans, depicting diverse body parts with distinct texture patterns. The images were captured by TissueScope LE 1.0 \footnote{http://www.hurondigitalpathology.com} 
bright field using a 0.75 NA lens. For each image, one can determine the resolution by checking the description tag in the header of the file. For instance, if the resolution is 0.5$\mu$m, then the magnification is 20x, and if the resolution is 0.25$ \mu $m, then the magnification is 40x. The dataset offers 27,055 training patches  and 1,325 (manually selected) test patches of size 1000$ \times $1000 (0.5mm$ \times $0.5mm) \cite{kimiapath24_2017}. The locations of the test patches in the scans have been removed (whitened) such that they cannot be mistakenly used for training. The color (\emph{staining}) is neglected in \kimiapath dataset; all patches are saved as grayscale images.  
The \kimiapath dataset is publicly available\footnote{http://kimia.uwaterloo.ca}.
\subsection{Patch Selection}
To create the \kimiapath dataset, each scan is divided into patches that are 1000$ \times $1000 pixels in size with no overlap between patches. Background pixels (i.e., very bright pixels) are set to \textit{white} and ignored using a homogeneity measure for each patch. The homogeneity for selection criterion is that every patch with a homogeneity of less than 99$ \% $ is ignored. The high threshold ascertains that no patch with significant texture pattern is ignored.  From the set of patches each scan had 100 randomly sampled patches are selected to be used for the fine-tuning process (we do not use all of them to emulate cases where no large dataset is available; besides more extensive training may destroy what a network has already learned). The values of each patch were subsequently normalized into $[0,1]$. The patches were finally downsized to a 224$ \times $224 to be fed into the CNN architecture.

\begin{figure*}[htb]
\centering
\vspace{0.08in}
\includegraphics[width=0.8\textwidth]{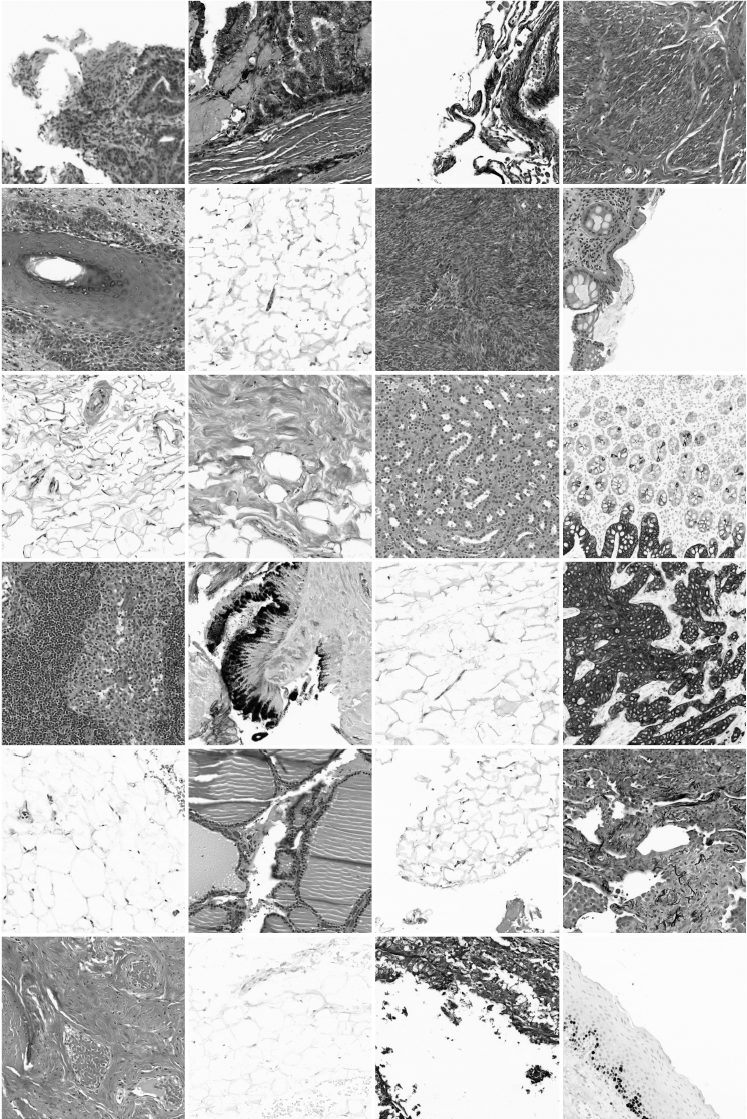}
\caption{A selection of patches from each training scan within the \kimiapath dataset. The patches are 1000$ \times $1000 pixels in size or 0.5mm$ \times $0.5mm. From top left to bottom right: scan/class 0 to scan/class 23.}
\label{fig:training-patches}
\end{figure*}

\begin{figure*}[tb]
\centering
\vspace{0.03in}
\includegraphics[width=0.47\textwidth]{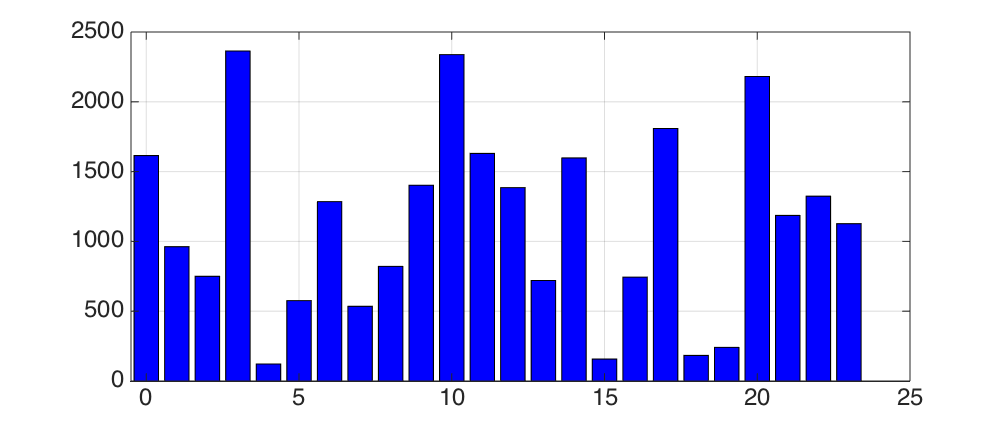}
\includegraphics[width=0.47\textwidth]{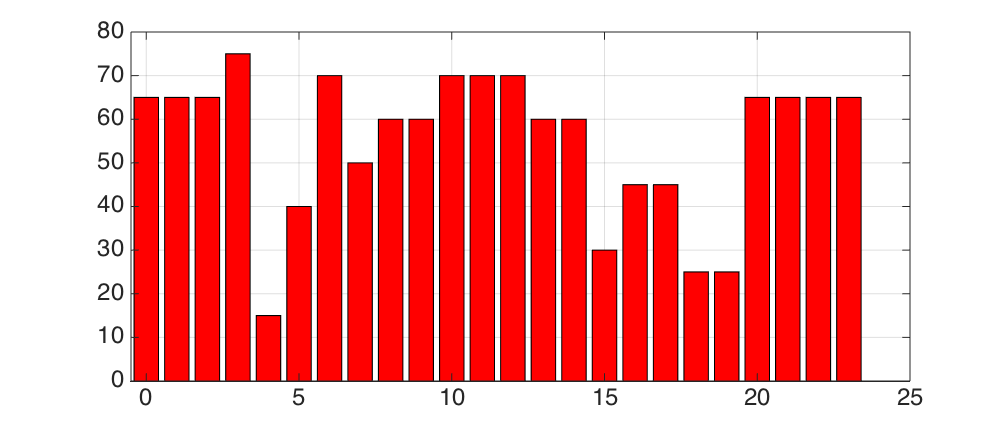}
\caption{Instance distribution for training set (left) and testing set (right) of \kimiapath.}
\label{fig:imbalance}
\end{figure*}

Following the above steps, we first obtained 27,055 patches from each scan based purely on the homogeneity threshold. Then, randomly sampled 100 patches from each class leading to the much smaller training set of 2,400 patches. A selection of patches from the training set can be viewed within Fig. \ref{fig:training-patches}. As Fig. \ref{fig:imbalance} shows the testing samples are relatively balanced in \kimiapath dataset, whereas the training set is rather imbalanced. Different size and frequency of specimens are the main reasons for the imbalance.

\subsection{Accuracy Calculation}
The accuracy measures used for the experiments are adopted from \cite{kimiapath24_2017}. These were chosen so that results between the papers could be compared. There are $ n_{tot} = 1,325 $ testing patches $ P_{s}^{j} $ that belong to 24 sets $ \Gamma_{s} = \{ P_{s}^{i} | s \in S, i = 1,2,\dots \} $ with $ s = 0,1,2,\dots,23 $ \cite{kimiapath24_2017}. Looking at the set of retrieved images for an experiment, $ R $, the \textbf{patch-to-scan accuracy}, $ \eta_{p} $, can be defined as

\begin{equation}
\label{patch2scan-accuracy}
\eta_{p} = \frac{1}{n_{tot}}\sum_{s \in S}|R \cap \Gamma_{s}|
\end{equation}

The \textbf{whole-scan accuracy}, $ \eta_{w} $, can be defined as

\begin{equation}
\label{wholescan-accuracy}
\eta_{w} = \frac{1}{24}\sum_{s \in S}|R \cap \Gamma_{s}|
\end{equation}

With the \textbf{total accuracy} is defined as $\eta_{total} = \eta_{p} \times \eta_{w}$.

\vspace{0.05in}
By incorporating both the accuracy measurements the resulting problem becomes much more difficult when attempting to obtain acceptable results \cite{kimiapath24_2017}.

\section{Methods}
Each experiment was run using the architecture for both the \emph{VGG16} and \emph{Inception-v3} networks as provided in the \emph{Keras Python} package \cite{VGG16, Inception-v3, keras-package}. Utilizing a pre-trained network, we then analyze the effectiveness of the network when using it just as a feature extractor, and when transferring the network (some of its weights) to the medical imaging domain. 

\subsection{Fine-Tuning Protocols}
When fine-tuning a deep network, the optimal setup varies between applications \cite{fine-tuning-colon}. However, using a pre-trained network and applying it to other domains has yielded better performing models \cite{cnn-transfer-learning}. It was decided that only the final convolutional block (block 5) within VGG16 and the final two inception blocks within Inception-v3 would be re-trained \cite{cnn-transfer-learning, fine-tuning-colon, VGG16, Inception-v3}. As in \cite{cnn-transfer-learning-training}  a single fully connected layer of size 256 (followed by an output layer of size 24) was chosen to replace the default VGG16 fully connected layers when fine-tuning. This was found to give better results. The optimizer we used follows the logic from \cite{cnn-transfer-learning, cnn-transfer-learning-training} where the learning rate chosen was very small ($ 10^{-4} $) and the momentum used was large ($ 0.9 $), both of which were selected to ensure no drastic changes within the weights of the network during training (which would destroy what had been already learned). The Keras data augmentation API was used to generate extra training samples and the network was trained for a total of 200 epochs (after which the accuracy was no longer changing) with a batch size of 32 \cite{keras-package}.

%
\subsection{Pre-Trained CNN as a Feature Extractor}
By using the provided implementation of the specified architectures within Keras, the pre-trained network was first used as a feature extractor without any fine-tuning (Feature Extractor VGG16 or FE-VGG16 and FE-Inception-v3)\cite{keras-package}. The last fully connected layer of the network -- prior to classification --  was used extracted to be used a feature vector. As pre-trained networks are trained in other domains (very different image categories) and hence cannot be used as classifier, we used the deep features to train a linear Support Vector Machine (SVM) for classification. The Python package \emph{scikit-learn} as well as LIBSVM were used to train SVM classifiers with a linear kernel \cite{scikit-learn,libsvm}. Both \emph{NumPy} and \emph{SciPy} were leveraged to manipulate and store data during these experiments \cite{numpy-scipy,numpy-scipy-package}.

%
%
%
%

\subsection{Fine-Tuned CNN as a Classifier}
\begin{figure}[!ht]
	\centering
	\includegraphics[width=0.96\columnwidth]{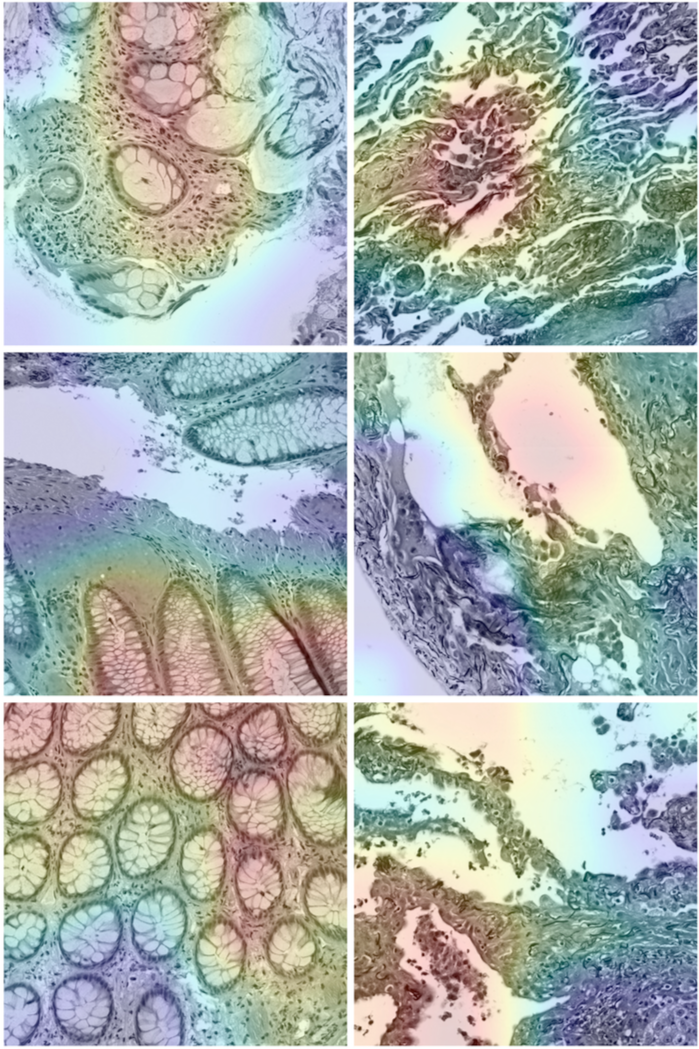}
	\caption{Activation maps using randomly selected patches from the \kimiapath testing data. The patches within each column are the same class and the labels per column are 4 and 8, respectively. The activation maps are created using the Keras Visualization Toolkit and the Grad-CAM algorithm \cite{keras-vis-package,grad-CAM,Zeiler2014}. Red areas had more influence on the label prediction \cite{grad-CAM}.}
	\label{fig:learned-kernels}
\end{figure}

The proposed network was then fine-tuned to the \kimiapath dataset. Using the Keras library, the convolutional layers were first separated from the top fully connected layers \cite{keras-package}. The training patches were fed through the model to create a set of \textit{bottleneck features} to initially pre-train the new fully-connected layers \cite{yu2011improved}. These features were used to initialize the weights of a fully connected MLP consisting of one 256 dense ReLU layer and a \emph{softmax} classification layer.  Next, the fully connected model was attached to the convolutional layers and training on each convolutional block, except the last block, was performed to adjust classification weights \cite{cnn-transfer-learning, cnn-transfer-learning-training}. Similarily, for the Inception-v3 network the fully connected layers were replaced with one 1024 dense ReLU layer and a \emph{softmax} classification layer. The fully connected layers were pretrained on \emph{bottleneck features} and then attached to the convolutional layers and training on the final two inception blocks was then performed. The resulting networks (Transfer Learned VGG16 or TL-VGG16 and TL-Inception-v3) were then used to classify the test patches. The class activation mappings (CAMs) for the fine-tuned Inception-v3 network on randomly selected test patches can be viewed in Fig. \ref{fig:learned-kernels}.

\section{Results}
The results of our experiments are summarized in Table \ref{tbl:SVMResults}. It can be stated the results for VGG16 and $ \text{CNN}_1 $are quite similar; training from scratch,  using a pre-trained network as feature extractor, and fine-tuning a pre-trained network are all delivering comparable results for \kimiapath. Whereas the results for Inception-v3 are similar with the transfer-learned model outperforming the feature extractor. As TL-Inception-v3 produced the best results, $ \eta_{total}=56.98\% $, and minimally updating the weights of a pre-trained network is not a time consuming task, one may prefer to utilize it. However, one may prefer using Inception-v3 to training from scratch and fine-tuning a pre-trained net as it requires no extra effort and produces similar results with a linear SVM.


\begin{table*}[!ht]
	\centering
	\vspace{0.08in}
	\caption{ Comparing the results training form scratch (CNN$_1$ reported in \cite{kimiapath24_2017}), using deep features via a pre-trained network with no change (FE-VGG16), and classification after fine-tuning a pre-trained network (TL-VGG16, TL-Inception-v3). The best scores are highlighted in bold.}
	\begin{tabular}{l|l|c|c|c}
		Scheme & Approach & $ \eta_{p} $ & $ \eta_{w} $ & $ \eta_{total} $ \\
		\hline
		Train from scratch & $ \text{CNN}_{1}$ \cite{kimiapath24_2017} & 64.98\% & 64.75\% & 41.80\% \\ \hline
		Pre-trained features & FE-VGG16 & 65.21\% & 64.96\% & 42.36\% \\
		Fine-tuning the pre-trained net & TL-VGG16 & 63.85\% & 66.23\% & 42.29\% \\ \hline
		Pre-trained features & FE-Inception-v3 & 70.94\% & 71.24\% & 50.54\% \\
		Fine-tuning the pre-trained net & TL-Inception-v3 & \textbf{74.87\%} & \textbf{76.10\%} & \textbf{56.98\%} \\ \hline
	\end{tabular}
	\label{tbl:SVMResults}
\end{table*}

\section{Discussions}
\begin{figure}[htbp]
	\centering
	\includegraphics[width=0.7\columnwidth]{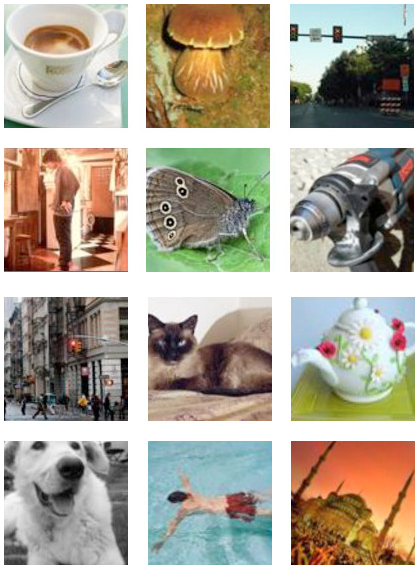}
	\caption{Sample images from \emph{ImageNet} project. One may object to using features that have been learned from such images in order to classify highly sensitive images of histopathology for medical diagnosis. However, experiments with \kimiapath dataset shows that features extracted from these images are expressive enough to compete against networks trained by histopathology images from scratch [Source: \emph{http://openai.com/}].  }
	\label{fig:imagenet}
\end{figure}

It was surprising to find out that simply using features from a pre-trained network (trained on non-medical images, see Fig. \ref{fig:imagenet}) can deliver results comparable with a network that, with considerable effort and resources, has been trained from scratch for the domain in focus (here histopathology). As well, such simpler approach was even able to achieve a noticeable accuracy increase of $ \approx 8.74\% $ in overall performance for \kimiapath dataset. Another surprising effect was that transfer learning via fine-tuning for VGG16 was not able to provide any improvement compared to extracting deep features from a pre-trained network without any change in the learned of its weights whereas with Inception-v3 the improvement was immediate. 

Perhaps the most obvious reaction to this finding is that if we had enough samples, i.e., millions of histopathological images, and if we would use proper computational devices for efficient training,  then CNN would perhaps deliver the best results clearly better than transfer learning. Although this statement is supported by comparable empirical evidence, it remains speculation for a sensitive field like medical imaging.

But why is so difficult to train a CNN for this case? It is most likely due to a number of factors such as a relative lack of image data, the effect of scaling down a patch for use within a deep network, architecture not well suited to the problem, or an overly simplistic fully connected network. However, as previously discussed in \cite{kimiapath24_2017}, the problem given by the \kimiapath dataset is indeed a hard problem, most likely due to the high variance between the different patches within a given scan (intra-class variability). This is further validated when looking at the results in Fig. \ref{fig:learned-kernels}. The two columns contain patches that have distinct patterns with their own unique features. The CAM from the first column shows that the network responds strongly to the unique structures within the 4 label (very strongly for the final patch). Whereas when presented with completely different patterns in the second column, the network responds strongly to other areas, typically ones that embody inner edges within the sample. This shows evidence that the model has at the very least begun to learn higher level structures within individual patches. Further investigation with different architectures would likely improve upon these results as would more aggressive augmentation. 

\section{Conclusions}
Retrieval and classification of histopathological images are useful but challenging tasks in analysis for diagnostic pathology. Whole scan imaging (WSI) generates gigapixel images that are immensely rich in details and exhibit tremendous inter- and intra-class variance.  
Both a feature extractor and transfer-learned network were able to offer  increases in classification accuracy on the \kimiapath dataset when compared to a CNN trained from scratch. Comparatively low performance of the latter could be due to the architecture not being well suited for the problem, lack of sufficient number of training images, and/or the inherent difficulty of the classification task for high-resolutional and highly variable histopathology images. Further work would warrant using different architectures for comparison, more aggressive data augmentation, and potentially increasing the size of training samples used from the \kimiapath dataset. However, both transfer-learned and feature extractor models were able to compete with the state-of-the-art methods reported in literature  \cite{kimiapath24_2017}, and therefore show potential for further improvements.

\section*{Acknowledgements}
The authors would like to thank \emph{Huron Digital Pathology} (Waterloo, ON, Canada) for its continuing support. 
 
\bibliographystyle{IEEEtran}
\bibliography{references}

\end{document}